\documentclass[sigconf]{acmart}

\AtBeginDocument{%
  \providecommand\BibTeX{{%
    \normalfont B\kern-0.5em{\scshape i\kern-0.25em b}\kern-0.8em\TeX}}}

\setcopyright{rightsretained}

\acmYear{2020}
\copyrightyear{2020}
\makeatletter
\renewcommand\@formatdoi[1]{\ignorespaces}
\makeatother
\acmISBN{}

\acmConference[KDD Converse'20]{KDD Workshop on Conversational Systems Towards Mainstream Adoption}{August 2020}

\usepackage{color}

\usepackage{graphicx}
\usepackage{adjustbox}
\usepackage{amsmath}
\DeclareMathOperator*{\argmax}{argmax}

\begin{document}

\title{A Fast and Robust BERT-based Dialogue State Tracker for Schema-Guided Dialogue Dataset}


\author{Vahid Noroozi}
\email{vnoroozi@nvidia.com}
\affiliation{%
    \institution{NVIDIA, USA}
}

\author{Yang Zhang}
\email{yangzhang@nvidia.com}
\affiliation{%
    \institution{NVIDIA, USA}
}

\author{Evelina Bakhturina}
\email{ebakhturina@nvidia.com}
\affiliation{%
    \institution{NVIDIA, USA}
}

\author{Tomasz Kornuta}
 \email{tkornuta@nvidia.com}
\affiliation{%
    \institution{NVIDIA, USA}
}

\renewcommand{\shortauthors}{Noroozi et al.}

\begin{abstract}
Dialog State Tracking (DST) is one of the most crucial modules for goal-oriented dialogue systems. In this paper, we introduce FastSGT (Fast Schema Guided Tracker), a fast and robust BERT-based model for state tracking in goal-oriented dialogue systems. The proposed model is designed for the Schema-Guided Dialogue (SGD) dataset which contains natural language descriptions for all the entities including user intents, services, and slots. The model incorporates two carry-over procedures for handling the extraction of the values not explicitly mentioned in the current user utterance. It also uses multi-head attention projections in some of the decoders to have a better modelling of the encoder outputs.

In the conducted experiments we compared FastSGT to the baseline model for the SGD dataset. Our model keeps the efficiency in terms of computational and memory consumption while improving the accuracy significantly. Additionally, we present ablation studies measuring the impact of different parts of the model on its performance. We also show the effectiveness of data augmentation for improving the accuracy without increasing the amount of computational resources.
\end{abstract}





\maketitle

\section{Introduction}
Goal-oriented dialogue systems is a category of dialogue systems designed to solve one or multiple specific goals or tasks (e.g. flight reservation, hotel reservation, food ordering, appointment scheduling) ~\cite{zhang2020recent}.  Traditionally, goal-oriented dialogue systems are set up as a pipeline with four main modules: 1-Natural Language Understanding (NLU), 2-Dialogue State Tracking (DST), 3-Dialog Policy Manager, and 4-Response Generator. NLU extracts the semantic information from each dialogue turn which includes e.g. user intents and slot values mentioned by user or system. DST takes the extracted entities to build the state of the user goal by aggregating and tracking the information across all turns of the dialogue. Dialog Policy Manager is responsible for deciding the next action of the system based on the current state. Finally, Response Generator converts the system action into human natural text understandable by the user.

The NLU and DST modules have shown to be successfully trained using data-driven approaches \cite{zhang2020recent}. In the most recent advances in language understanding, due to models like BERT \cite{devlin2019bert}, researchers have successfully combined NLU and DST into a single unified module, called Word-Level Dialog State Tracking (WL-DST)~\cite{wu2019transferable, kim2019efficient, rastogi2019towards}. The WL-DST models can take the user or system utterances in natural language format as input and predict the state at each turn. The model we are going to propose in this paper falls into this class of algorithms.  

Most of the previously published public datasets, such as MultiWOZ \cite{budzianowski2018multiwoz} or M2M \cite{shah2018building}, use a fixed list of defined slots for each domain without any information on the semantics of the slots and other entities in the dataset. As a result, the systems developed on these datasets fail to understand the semantic similarity between the domains and slots. The capability of sharing the knowledge between the slots and domains might help a model to work across multiple domains and/or services, as well as to handle the unseen slots and APIs when the new APIs and slots are similar in functionality to those present in the training data.

The Schema-Guided Dialogue (SGD) dataset \cite{rastogi2019towards} was created to overcome these challenges by defining and including schemas for the services. A \texttt{schema} can be interpreted as an ontology encompassing naming and definition of the entities, properties and relations between the concepts. In other words, \texttt{schema} defines not only the structure of the underlying data (relations between all the services, slots, intents and values), but also provides descriptions of most of the entities expressed in a natural language. As a result, the dialogue systems can exploit that rich information to capture more general semantic meanings of the concepts. Additionally, the availability of the schema enables the model to use the power of pre-trained models like BERT to transfer or share the knowledge between different services and domains. The recent emergence of the SGD dataset has triggered a new line of research on dialogue systems based on schemas, e.g.~\cite{balaraman2020domain,ruan2020fine,lei2020zero,gulyaev2020goal}.

Many state-of-the-art models proposed for the SGD dataset, despite showing impressive performance in terms of accuracy, appear not to be very efficient in terms of computational complexity and memory consumption, e.g.~\cite{ma2019end,miaoli2020,xingshi2020,lei2020zero}.

To address these issues, we introduce a fast and flexible WL-DST model called Fast Schema Guided Tracker (FastSGT)\footnote{Source code of the model is publicly available at: https://github.com/NVIDIA/NeMo}. Main contributions of the paper are as follows:
\begin{itemize}
    \item FastSGT is able to predict the whole state at each turn with just a single pass through the model, which lowers both the training and inference time.
    \item The model employs carry-over mechanisms for transferring the values between slots, enabling switching between services and accepting the values offered by the system during dialogue.
    \item We propose an attention-based projection to attend over all the tokens of the main encoder to be able to model the encoded utterances better.
    \item We evaluate the model on the SGD dataset~\cite{rastogi2019towards} and show that our model has significantly higher accuracy when compared to the baseline model of SGD, at the same time keeping the efficiency in terms of computation and memory utilization.
    \item We show the effectiveness of augmentation on the SGD dataset without increasing the training steps.

\end{itemize}

\section{Related Works}
\label{prev_works}
The availability of schema descriptions for services, intents and slots enables the NLU/DST models to share and transfer knowledge between different services that have similar slots and intents. Considering the recent advances in natural language understanding and rise of Transformer-based models~\cite{vaswani2017attention} like BERT~\cite{devlin2019bert} or RoBERTa~\cite{liu2019roberta}, it looks like a promising approach for training a unified model on datasets which are aggregated from different sources.
We categorize all models proposed for the SGD dataset into two main categories: multi-pass and single-pass models.

\subsection{Multi-pass Models}
The general principle of operation of multi-pass models~\cite{xingshi2020,ma2019end,miaoli2020,junyuan2020,ruan2020fine} lies in passing of descriptions of every slots and intents as inputs to the BERT-like encoders to produce their embeddings. As a result encoders are executed several times per a single dialog turn. Passing the descriptions to the model along with the user or system utterances enables the model to have a better understanding of the task and facilitates learning of similarity between intents and slots. The SPDD model~\cite{miaoli2020} is a multi-pass model which showed one of the highest performances in terms of accuracy on the SGD dataset. For instance, in order to predict the user state for a service with 4 intents and 10 slots and 3 slots being active in a given turn, this model needs 27 passes through the encoder (4 for intents, 10 for requested status, 10 for statuses, and 3 for values). 
Such approaches handle unseen services well and achieve high accuracy, but seem not to be practical in many cases when time or resources are limited.

One obvious disadvantage of multi-pass models is their lack of efficiency.
The other disadvantage is the memory consumption. They typically use multiple BERT-like models (e.g. five in SPDD) for predicting intents, requested slots, slot statuses, categorical values, and non-categorical values. This significantly increases the memory consumption compared to most of the single-pass models with a single encoder.

\subsection{Single-pass Models}
The works that incorporate the single-pass approach \cite{rastogi2019towards,balaraman2020domain} rely on BERT-like models to encode the descriptions of services, slots, intents and slot values into representations, called \texttt{schema embeddings}.
The main difference lies in the fact that this procedure is executed just once, before the actual training starts, mitigating the need to pass the descriptions through the model for each one of the turns/predictions. 

While these models are very efficient and robust in terms of training and inference time, they have shown significantly lower performance in terms of accuracy compared to the multi-pass approaches. On the other hand, multi-pass models need significantly higher computation resource for training and inference, and also the usage of additional BERT-based encoders increases the memory usage drastically.

\section{The FastSGT Model}
\label{fast_sgd_model}

\begin{figure*}[t!]
\centering
\includegraphics[width=0.9\textwidth]{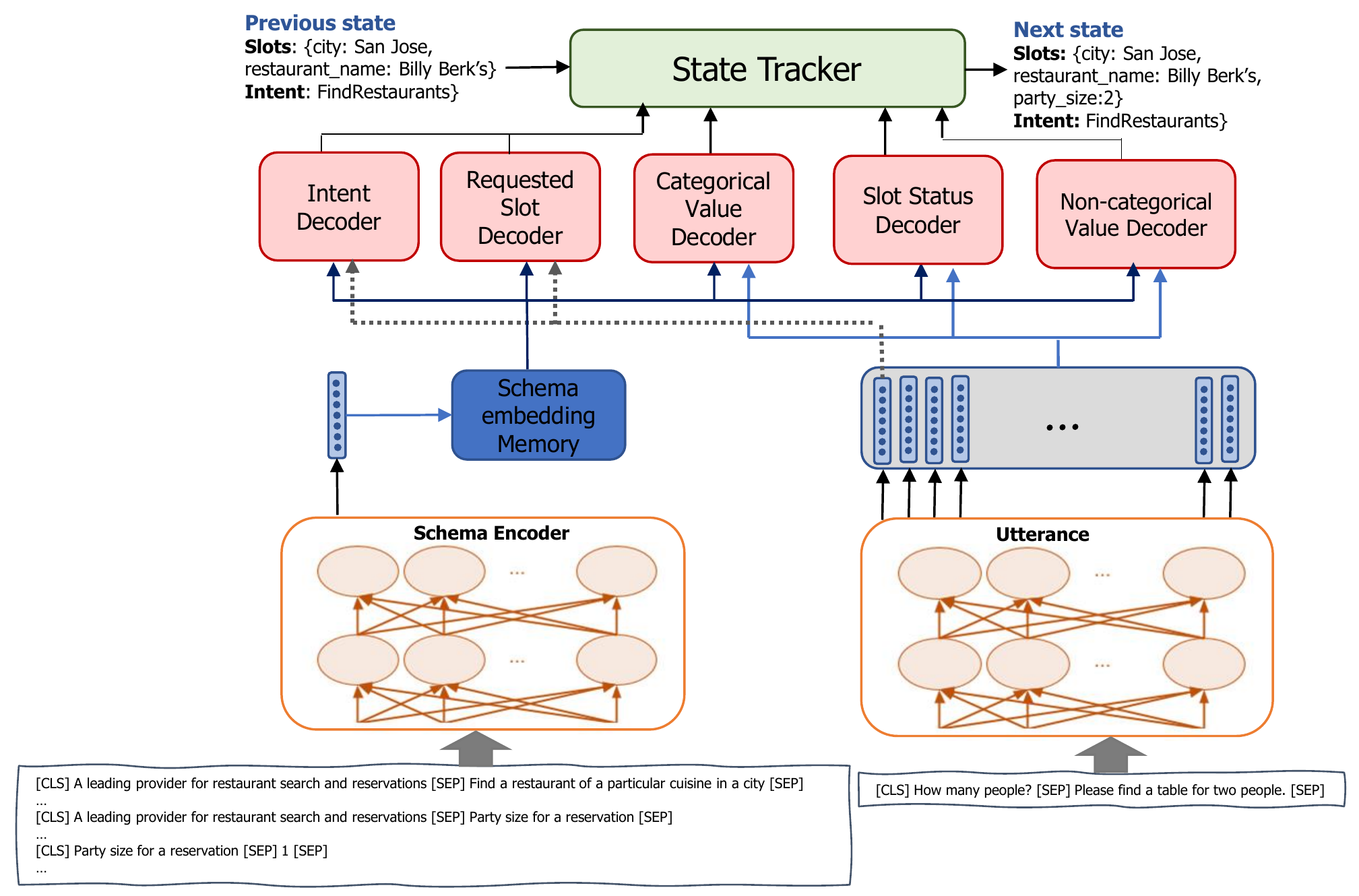}
\caption{The overall architecture of FastSGT (Fast Schema Guided Tracker) with exemplary inputs from a restaurant service.}
\label{fig:model_arch}
\end{figure*}

The FastSGT (Fast Schema Guided Tracker) model belongs to the category of single-pass models, keeping the flexibility along with memory and computational efficiency. Our model is based on the baseline model proposed for SGD~\cite{rastogi2019towards} with some improvements in the decoding modules.

The model architecture is illustrated in Fig.~\ref{fig:model_arch}. It consists of four main modules: 1-\textbf{Utterance Encoder}, 2-\textbf{Schema Encoder}, 3-\textbf{State Decoder}, and 4-\textbf{State Tracker}. The first three modules constitute the NLU component and are based on neural networks, whereas the state tracker is a rule-based module.
We used BERT~\cite{devlin2019bert} for both encoders in our model, but similar models like RoBERTa \cite{liu2019roberta} or XLNet~\cite{yang2019xlnet} can also be used. 

Assume we have a dialogue of $N$ turns. Each turn consists of the preceding system utterance ($S_t$) and the user utterance ($U_t$). Let $D=\{(S_1, U_1), (S_1, U_2), ..., (S_N, U_N)\}$ be the collection of turns in the dialogue. 

The \textbf{Utterance Encoder} is a BERT model which encodes the user and system utterances at each turn. The \textbf{Schema Encoder} is also a BERT model which encodes the schema descriptions of intents, slots, and values into schema embeddings. These schema embeddings help the decoders to transfer or share knowledge between different services by having some language understanding of each slot, intent, or value.
The schema and utterance embeddings are passed to the \textbf{State Decoder} - a multi-task module. This module consists of five sub-modules producing the information necessary to track the state of the dialogue.
Finally, the \textbf{State Tracker} module takes the previous state along with the current outputs of the \textbf{State Decoder} and predicts the current state of the dialogue by aggregating and summarizing the information across turns.
Details of all model components are presented in the following subsections.

\subsection{Utterance Encoder}
This module is responsible for encoding of the current turn of the dialogue. At each turn $(S_{t}, U_t)$, the preceding system utterance is concatenated with the user utterance separated by the special token of $[SEP]$, resulting in ($T_t$) which is serves as input into the utterance encoder module:

\begin{equation}
T_t = [CLS] \oplus  S_{t} \oplus [SEP] \oplus U_t \oplus [SEP]
\end{equation}

The output of the first token passed to the encoder is denoted as $Y_{cls}$ and is interpreted as a sentence-level representation of the turn, whereas the token-level representations are denoted as $Y_{tok} = [Y_{tok}^1, Y_{tok}^2, ..., Y_{tok}^M]$, where M is the total number of tokens in $T_t$. 

\subsection{Schema Encoder}
The \textbf{Schema Encoder} uses the descriptions of intents, slots, and services to produce some embedding vectors which represent the semantics of slots, intents and slot values. To build these schema representations we instantiate a BERT model with the same weights as the \textbf{Utterance Encoder}. However, this module is used just once, before the training starts, and all the schema embeddings are stored in a memory to be reused during training. This means, they will be fixed during the training time. This approach of handling the schema embeddings is one of the main reasons behind the efficiency of our model compared to the multi-pass models in terms of computation time.

We used the same approach introduced in \cite{rastogi2019towards} for encoding the schemas. For a service with $N_{I}$ intents, $N_{C}$ categorical slots and $N_{NC}$ non-categorical slots, the representation of the intents are denoted as $I_i$, $ 1 \leqslant i \leqslant N_I$. Schema embeddings for the categorical and non-categorical slots are indicated as $S^{C}_i$, $ 1 \leqslant i \leqslant N_{C}$, and $S^{NC}_i$, $ 1 \leqslant i \leqslant N_{NC}$ respectively. The embeddings for the values of the $k$-th categorical slot of a service with $N_{V}^k$ possible values is denoted as $V_i^k, 1 \leqslant i \leqslant N_{V}^k$.

Generally, the input to the \textbf{Schema Encoder} is the concatenation of two sequences with the $[SEP]$ token used as the separator and the $[CLS]$ token indicating the beginning of the sequence.

The \textbf{Schema Encoder} produces four types of schema embeddings:
intents, categorical slots, non-categorical slots and categorical slot values. For a single intent embeddings $I_i$, the first sequence is the corresponding service description and second one is the intent description. For each categorical $S^{C}_i$ and non-categorical $S^{NC}_i$ slots embedding, the service description is concatenated with the description of the slot.
To produce the schema embedding $N_{V}^k$ for the k-$th$ possible value of a categorical slot, the description of the slot is used as the first sequence along with the value itself as the second sequence.

These sequences are given one by one to the \textbf{Schema Encoder} before the main training is started and the output of the first output token embedding $Y_{cls}$ is extracted and stored as the schema representation, forming the \textbf{Schema Embeddings Memory}.

\subsection{State Decoder}
The \textbf{Schema Embeddings Memory} along with the outputs of the \textbf{Utterance Encoder} are used as inputs to the \textbf{State Decoder} to predict the values necessary for state tracking.
The \textbf{State Decoder} module consists of five sub-modules, each employing a set of projection transformations to decode their inputs.
We use the two following projection layers in the decoder sub-modules:

1)~~\texttt{Single-token projection}: this projection transformation, which is introduced in \cite{rastogi2019towards}, takes the schema embedding vector and the $Y_{cls}$ of the \textbf{Utterance Encoder} as its inputs. The projection for predicting $p$ outputs for task $K$ is defined as $F^{K}_{FC}(x,y;p)$ for two vectors $x, y \in R^q$ as the inputs. $q$ is the embedding size, $p$ is the size of the output (e.g. number of classes), the first input $x$ is a schema embedding vector, and $y$ is the sentence-level output embedding vector produced by the \textbf{Utterance encoder}. The sources of the inputs $x$ and $y$ depend on the task and the sub-module. Function $F^{K}_{FC}(x,y;p)$ for projection $K$ is defined as:

\begin{equation}
    h_1 = GELU(W_1^Ky+b_1^K)
\end{equation}
\begin{equation}
    h_2 = GELU(W_2^K(x \oplus h_1) + b_2^K)
\end{equation}
\begin{equation}
    F^{K}_{FC}(x,y;p) = Softmax(W_3^Kh_2 +b_3^K)
\end{equation}

\noindent 
where $W_i^K, 1 \leqslant i \leqslant 3$ and $b^K_i, 1 \leqslant i \leqslant 3$ are the learnable parameters for the projection, and $GELU$ is the activation function introduced in \cite{hendrycks2016gaussian}. Symbol $\oplus$ indicates the concatenation of two vectors. Softmax function is used to normalize the outputs as a distribution over the targets. This projection is used by the \textbf{Intent}, \textbf{Requested Slot} and \textbf{Non-categorical Value Decoders}.

2)~~\texttt{Attention-based projection}: the single-token projection just takes one vector from the outputs of the \textbf{Utterance Encoder}. For the \textbf{Slot Status Decoder} and \textbf{Categorical Value Decoder} we propose to use a more powerful projection layer based on multi-head attention mechanism~\cite{vaswani2017attention}.
We use the schema embedding vector $x$ as the query to attend to the token representations $Y_{tok}$ as outputted by the \textbf{Utterance Encoder}. The idea is that domain-specific and slot-specific information can be extracted more efficiently from the collection of token-level representations than just from the sentence-level encoded vector $Y_{cls}$. The multi-head attention-based projection function $F^{K}_{MHA}(x,Y_{tok};p)$ for task $K$ to produce targets with size $p$ is defined as:

\begin{equation}
    h_1 = \text{MultiHeadAtt}(query=x, keys=Y_{tok}, values=Y_{tok})
\end{equation}

\begin{equation}
    F^{K}_{MHA}(x,Y_{tok};p) = Softmax(W_1^Kh_1 + b_1^K)
\end{equation}
\noindent 
where $MultiHeadAtt$ is the multi-head attention function introduced in~\cite{vaswani2017attention}, and $W_1^K$ and $b^K_1$ are learnable parameters of a linear projection after the multi-head attention. To accommodate padded utterances we use attention masking to mask out the padded portion of the utterance.

\subsubsection{\textbf{Intent Decoder}}
Each service schema contains a list of all possible user intents for the current dialogue turn. For example, in a service for reserving flight tickets, we may have intents of searching for a flight or cancelling a ticket. At each turn, for each service, at most one intent can be active. For all services, we additionally consider an intent called $NONE$ which indicates that there is no active intent for the current turn. An embedding vector $I_0$ is considered as the schema embedding for the $NONE$ intent. It is a learnable embedding which is shared among all the services.

The inputs to the Intent Decoder for a service are the schema embeddings $I_i, 0 \leqslant i \leqslant N_I$ from the \textbf{Schema Embeddings Memory} and $Y_{cls}$ of the \textbf{Utterance Encoder}. The predicted output of this sub-module is the active intent of the current turn $I_{active}$ defined as:

\begin{equation}
    I_{active} = \argmax_{0 \leqslant i \leqslant N_I} F^{K}_{FC}(I_i, Y_{cls}; p=N_I)
\end{equation}

\subsubsection{\textbf{Slot Request Decoder}}
At each turn, the user may request information about a slot instead of informing the system about a slot value. For example, when a user asks for the flight time when using a ticket reservation service, $flight\_time$ slot of the service is requested by the user. This is a binary prediction task for each slot. For this task, slot $s_i$ is requested when $F^{K}_{FC}(S^i_j, Y_{cls};p=1) > 0.5, j \in {c, n}$. The same prediction is done for both categorical and non-categorical slots.

\subsubsection{\textbf{Slot Status Decoder}}
We consider four different statuses for each slot, namely: $inactive, active, dont\_care, carry\_over$. If the value of a slot has not changed from the previous state to the current user state, then the slot's status is "$inactive$". If a slot's value is updated in the current user's state into "$dont\_care$", then the status of the slot is set to "$dont\_care$" which means the user does not care about the value of this slot. If the value for the slot is updated and its value is mentioned in the current user utterance, then its status is "$active$". There are many cases where the value for a slot does not exist in the last user utterance and it comes from previous utterances in the dialogue. For such cases, the status is set to "$carry\_over$" which means we should search the previous system or user utterances in the dialogue to find the value for this slot. More details of the carry over mechanisms are described in subsection \ref{state_tracker}.

The status of the categorical slot $r$ is defined as:

\begin{equation}
    S_r = \argmax_{0 \leqslant i \leqslant N_c} F^{K}_{MHA}(S_i, Y_{tok}; p=4)
\end{equation}

Similar decoder is used for the status of the non-categorical slots as:

\begin{equation}
    S_r = \argmax_{0 \leqslant i \leqslant N_{nc}} F^{K}_{MHA}(S_i, Y_{tok}; p=4)
\end{equation}

\noindent 
where $S_r$ is the status of $r$-th non-categorical slot.

\subsubsection{\textbf{Categorical Value Decoder}}
The list of possible values for categorical slots are fixed and known. For the active service, this sub-module takes the $Y_{cls}$ and the schema embeddings of the values for all the categorical slots $V_i^k, 1 \leqslant k \leqslant N_c, 1 \leqslant i \leqslant N_V^k$ as input and uses the $F^{K}_{MHA}(x,Y_{tok};p=1)$ projection on each value embedding and $Y_{tok}$. Then the predictions of the values are normalized by another Softmax layer, resulting in a distribution over all possible values for each slot. The value with the maximum probability is selected as the value of the slot. If a slot's status is not "$active$" or "$carry\_over$", this prediction for a given slot will be ignored during the training.

\subsubsection{\textbf{Non-categorical Value Decoder}}
For non-categorical slots there is no predefined list of possible values, and they can take any value. For such slots, we use the spanning network approach~\cite{rastogi2019towards} to decode and extract the slot's value directly from the user or system utterances. 

This decoder would use two projection layers, one for predicting the start of the span, and one for end of the span. This projection layer is shared among all the non-categorical slots. The only difference is their inputs. For each non-categorical slot, the outputs of the tokens from the \textbf{Utterance Encoder} and also the schema embedding for the slot are given as inputs to the two projection layers. These predict the probability of each token to be the start or end of the span for the value. The token with the maximum probability of being the start token is considered as the start and the token with the maximum probability of being the end token which appears after the start token is be considered as the end of the span. 

For slots with no value mentioned in the current turn, we train the network to predict the first $[CLS]$ to be both the start and end tokens. The non-categorical slots with status of "$inactive$" are ignored during the training.

\subsection{State Tracker}
\label{state_tracker}
The user's state at each turn includes the active intent, the collection of the slots and their corresponding values, and also the list of the requested slots. This module would get all the predictions from the decoders along with the previous user state and produces the current user's state. The list of requested slots and the active intent are produced from the outputs of the \textbf{Slot Request Decoder} and \textbf{Intent Decoder} directly. Producing the collection of the slots and their values requires more processing.

We start with the list of slots and values from the previous state, and update it as the following: For each slot, if its status is "$inactive$", no update or change is applied and the value from the previous state is be carried over. If its status is predicted as "$dont\_care$", then the explicit value of "don't care" would be set as the value for the slot. If the slot status is "$active$", then it means we should retrieve a value for that slot. We initially check the output of the value decoders, if a value is predicted, then that value is assigned to the slot. If the "$\#CARRYOVER\#$" special value is predicted for categorical slots or the start and end pointers for the non-categorical slots are pointing to a value outside the user utterance's span, then we kick off the carry-over mechanism. The carry-over mechanisms try to search the previous states or system actions for a value. When the "$carry\_over$" status is predicted for any slot, the same carry-over procedures are triggered.

The carry-over procedures~\cite{miaoli2020} enable the model to retrieve a value for a slot from the preceding system utterance or even previous turns in the dialogue. There are many cases where the user is accepting some values offered by the system and the value is not mentioned explicitly in the user utterance. 
In our system we have implemented two different carry-over procedures. The value may be offered in the last system utterance, or even in the previous turns by the system. The procedure to retrieve values in these cases is called in-service carry-over. 
There are also cases where a switch is happening between two services in multi-domain dialogues. A dialogue may contain more than one service and the user may switch between these services. When a switch happens, we may need to carry some values from a slot in the previous service to another slot in the current service. This carry-over procedure is called cross-service carry-over. 
\subsubsection{\textbf{In-service Carry-over}}
We trigger this procedure in three cases: 1-status of a slot is predicted as "$carry\_over$", 2-the spanning region found for the non-categorical slots is not in the span of the user utterance, 3-"$\#CARRYOVER\#$" value is predicted for a categorical slot with "active" or "$carry\_over$" statuses. The in-service carry-over procedure tries to retrieve a value for a slot from the previous system utterances in the dialogue. We first search the system actions starting from the most recent system utterance and then move backwards for a value mentioned for that slot. The most recent value would be considered for the slot if multiple values are found. If no value could be found, that slot would not get updated in the current state.
\subsubsection{\textbf{Cross-service Carry-over}}
Carrying values from previous services to the current one when a switch happens in a turn is done by cross-service carry-over procedure.

The previous service and slots are called sources, and the new service and slots are called the targets. To perform the carry-over, we need to build a list of candidates for each slot which contains the slots where a carry-over can happen from them. We create this carry-over candidate list from the training data. We process the whole dialogues in the training data, and count the number of times a value from a source slot and service carry-overs to a target service and slot when a switch happens. These counts are normalized to the number of the switches between each two services in the whole training dialogues to have a better estimate of the likelihood of carry-overs. In our experiments, the candidates with likelihoods less than $0.1$ are ignored. This carry-over relation between two slots is considered symmetric and statistics from both sides are aggregated. This candidate list for each slot contains a list of slot candidates from other services which is looked up to find a carry-over value.

When a switch between two services happens in the dialogue, the cross-service carry-over procedure is triggered. It looks into the candidate for all the slots of the new service and carry any value it found from previous service. If multiple values for a slot are found in the dialogue, the most recent one is used. 

\subsubsection{\textbf{State Summarization}}
At each turn, the predictions of the decoders are used to update the previous state and produce the current state. The slots and their values from the previous state are updated with the new predictions if their statuses are predicted as active, otherwise kept unchanged. In case of several WL-DST models, e.g. TRADE~\cite{wu2019transferable} or the the model introduced in \cite{xingshi2020}, they require the entire of part of the dialogue history to be passed as the input. FastSGT just decodes the current turn and updates just the parts of the state that need the change at each turn. It helps the computational efficiency and robustness of the model.

\section{Experiments}
\label{exps}

\subsection{Experimental Settings}
We did all our experiments on systems with 8 V100 GPUs using mixed precision training \cite{micikevicius2017mixed} to make the training process faster. We used BERT-base-cased in our experiments for the encoders. All of the models were trained for maximally $160$ epochs to have less variance in the results, whereas most of them managed to converge in less than $60$ epochs. We repeated each experiment three times and reported the average in all tables. 

For each of the BERT-based encoders and attention-based projection layers we used $16$ heads. We have optimized the model using Adam~\cite{kingma2014adam} optimizer with default parameter settings. Batch size was set to 128 per GPU, with maximum learning rate to $4e-4$. Linear decay annealing was used with warm-up of $2\%$ of the total steps. Dropout rate of $0.2$ is used for regularization for all the modules. We tuned all the parameters to get the best joint goal accuracy on the \textit{dev}-set of the datasets.

\subsection{Datasets}
The SGD dataset is multi-domain goal-oriented dataset that contains over 16k multi-domain conversations between a human user and a virtual assistant and spans across 16 domains. The SGD dataset provides schemas that contain details of the API’s interface. The schema define the list of the supported slots by each service, possible values for categorical slots, along with the supported intents.

In the original SGD dataset, test and validation splits contains \textit{Unseen Services}, i.e. services that do not exist in the training set.
By design, those services are in most cases similar to the services in the train with similar descriptions, but different names.
We created another version of the dataset by merging all the dialogues and spitting them again randomly in $70\%/15\%/15\%$ to train/validation/test.
In this version of the dataset, called $SGD+$, all services are seen.
It helped us to have a better evaluation of our model as the focus of our model is for seen services and by using the original split we were ignoring a significant part of the dataset (some services simply weren't present in the original \textit{Seen Services} split).

\subsection{Evaluation Metrics}
\label{metrics}
We evaluated our proposed model and compared with other baselines on the the following metrics:

\begin{itemize}
  \item \textbf{Active Intent Accuracy}: the fraction of user turns with correctly predicted active intent, where active intent represents user’s intent in the current turn that is being fulfilled by the system.
  \item \textbf{Requested Slot F1}: the macro-averaged F1 score of the requested slots. The turns with no requested slots are skipped. Requested slots are the slots that were requested by the user in the most recent utterance.
  \item \textbf{Average Goal Accuracy}: the average accuracy of predicting the value of a slot correctly, where the user goal represents user’s constraints mentioned in the dialogue up until the current user turn.
  \item \textbf{Joint Goal Accuracy}: the average accuracy of predicting all slot values in a dialogue turn correctly. This metric is the strictest one among them all.
\end{itemize}

\begin{table*}[ht]
\caption{\label{final-evaluation1}Evaluation results of our model compared to the SGD baseline \cite{rastogi2019towards} on the dev/test sets of the SGD dataset. The results are reported separately for single-domain dialogues and all dialogues. We also reported the results without the fixes in the evaluation process of the baseline for both seen and all services.}
\centering
\begin{adjustbox}{max width=\textwidth}
\begin{tabular}{l|cc|cccc|cccc}
 & & \textbf{Eval} & \multicolumn{4}{c|}{Single-domain Dialogues}  & \multicolumn{4}{c}{All dialogues} \\
 \textbf{Model} & \textbf{Services} & \textbf{Fix} & \textbf{Intent Acc} & \textbf{Slot Req F1}& \textbf{Average GA} & \textbf{Joint GA} & \textbf{Intent Acc} & \textbf{Slot Req F1} & \textbf{Average GA} & \textbf{Joint GA} \\ \hline

SGD Baseline & all & --  
& 96.00/88.05 & 96.50/95.60 & 77.60/68.40 & 48.60/35.60 
& 90.80/90.60 & 97.30/96.50 & 74.00/56.00 & 41.10/25.40 \\

FastSGT & all & --
& 96.45/88.60 & 96.55/94.65 & 81.11/71.22 & \textbf{56.66/39.77} 
& 91.58/90.33 & 97.53/96.33 & 78.22/60.66 & \textbf{52.06/29.20} \\  \hline

SGD Baseline & seen & --  
& 99.03/78.22 & 98.74/96.83 & 88.12/92.17 & 68.61/73.94 
& 96.44/94.50 & 99.47/99.29 & 79.86/67.77 & 54.68/41.63 \\

FastSGT & seen & --
& 98.94/77.53 & 98.80/96.89 & 92.98/94.12 & \textbf{83.13/80.25}
& 96.61/94.18 & 99.66/99.55 & 88.78/76.52 & \textbf{71.34/55.23} \\ \hline

SGD Baseline & seen & +
& 99.00/75.12 & 96.08/99.22 & 90.84/91.42 & 71.14/68.94 
& 96.08/91.64 & 99.62/99.66 & 83.33/81.03 & 61.15/60.05 \\

FastSGT & seen & + & 98.86/73.98 & 99.64/99.24 & 96.54/95.31 & \textbf{88.03/81.56}
& 96.26/91.44 & 99.65/99.64 & 92.33/92.12 & \textbf{79.65/78.55} \\ 

\end{tabular}
\end{adjustbox}
\end{table*}

\begin{table*}[ht]
\caption{\label{final-evaluation2} Evaluation results of our model compared to the SGD baseline \cite{rastogi2019towards} on SGD+ dataset for single-domain dialogues and also all dialogues. All dialogues includes all the single-domain and multi-domain dialogues.}

\centering
\begin{adjustbox}{max width=0.95\textwidth}
\begin{tabular}{l|cccc|cccc}
 & \multicolumn{4}{c|}{Single-domain Dialogues}  & \multicolumn{4}{c}{All Dialogues} \\
 \textbf{Model}  & \textbf{Intent Acc} & \textbf{Slot Req F1} & \textbf{Average GA} & \textbf{Joint GA} & \textbf{Intent Acc} & \textbf{Slot Req F1} & \textbf{Average GA} & \textbf{Joint GA} \\
\hline
SGD Baseline & 95.31/95.70 & 99.49/99.66 & 91.85/91.67 & 71.76/71.86 & 96.34/96.45 & 99.69/99.69 & 81.25/81.15 & 56.07/56.11 \\
 
FastSGT & 95.36/95.67 & 99.52/99.59 & 95.57/95.44 & \textbf{83.97/84.31} & 96.17/96.36 & 99.69/99.69 & 95.33/95.46 & \textbf{82.03/82.81} \\ 
\end{tabular}
\end{adjustbox}
\end{table*}

\subsection{Performance Evaluation}
In this section, we evaluate and compare FastSGT with the baseline SGD model proposed in \cite{rastogi2019towards} on the two datasets of $SGD$ and $SGD+$. A through comparison of both models with regard to all the evaluation metrics explained in subsection \ref{metrics} are reported in Tables \ref{final-evaluation1} and \ref{final-evaluation2}. We have reported the metrics calculated for the turns from the all services vs just seen services. The performance of the models trained and evaluated on single-domain and all-domains are reported separately. All dialogues includes single-domain and multi-domain dialogues.

There were some issues in the original evaluation process of the SGD baseline which we had to fix. First, some services were considered seen services during the evaluation for single-domain dialogues while they do not actually exist in the training data. The other issue was that the turns which come after an unseen service in multi-domain dialogues could be counted as seen by the original evaluation. The errors from unseen services may propagate through the dialogue and affect some of the metrics for seen services. We fixed it by just considering only the turns if there are no turns before them in the dialogue from unseen services. These fixes helped to improve the results. However, to have a fair comparison we also reported the performance of the baseline model and ours with and without these fixes. In Table \ref{final-evaluation1}, we have denoted them by +/- in the column labelled as \textbf{"Eval Fix"}.

The performance of our model in all the metrics on both versions of the dataset is better than the baseline. The main advantage of our model comparing to the baseline model is benefiting from the carry-over mechanisms. These procedures are enhanced by having the capability of predicting "carryover" statuses for all slots, and also having "\#CARRYOVER\#" value for the categorical slots. During our error analysis, we found out that great number of errors come from the cases where the value for a slot is not explicitly mentioned in the user utterance. The other advantage of FastSGT over the SGD baseline is the attention-based projection functions which enable our model to have a better modeling of the utterance encoder's outputs.

\begin{table*}[ht]
\caption{\label{ablation-study} The results of the ablation study of FastSGT on test/dev sets of SGD+. }

\centering
\begin{adjustbox}{max width=0.95\textwidth}
\begin{tabular}{l|cccc|cccc}
 & \multicolumn{4}{c|}{Single-domain Dialogues}  & \multicolumn{4}{c}{All Dialogues} \\
 \textbf{Model}  & \textbf{Intent Acc} & \textbf{Slot Req F1} & \textbf{Average GA} & \textbf{Joint GA} & \textbf{Intent Acc} & \textbf{Slot Req F1} & \textbf{Average GA} & \textbf{Joint GA} \\ 
 \hline
 SGD Baseline 
 & 95.31/95.70 & 99.49/99.66 & 91.85/91.67 & 71.76/71.86 
 & 96.34/96.45 & 99.69/99.69 & 81.25/81.15 & 56.07/56.11 \\ 

 \hline
 FastSGT 
 & 95.36/95.67 & 99.52/99.59 & 95.57/95.44 & \textbf{83.97/84.31} 
 & 96.17/96.36 & 99.69/99.69 & 95.33/95.46 & \textbf{82.03/82.81}\\

 - Attention-based Layer & 95.23/95.72 & 99.51/99.62 & 95.28/95.04 & 82.85/83.04 
 & 96.32/96.48 & 99.69/99.70 & 95.14/95.44 & 81.80/82.76 \\

   - Carry-over Value/Status  
   & 95.12/95.87 & 99.54/99.63 & 95.30/95.40 & 81.88/82.36
   & 96.32/96.50 & 99.68/99.71 & 95.08/95.27 & 80.84/81.65 \\

    - In-service Carry-over & 95.28/95.83 & 99.50/99.65 & 91.48/91.18 & 71.48/71.34 
    & 96.25/96.41 & 99.67/99.70 & 92.07/92.01 & 72.62/72.58 \\ 

    - Cross-service Carry-over & - & - & - & - 
    & 96.28/96.43 & 99.68/99.71 & 84.31/84.37 & 62.90/63.54 \\  
 
\end{tabular}
\end{adjustbox}
\end{table*}

\subsection{Data Augmentation}
One of the main challenges in building goal-oriented dialogue system is lack of sufficient high quality data with annotation. In this section, we study the effectiveness of augmentation on the performance of our model. We created augmented versions of the SGD and SGD+ datasets by replacing the values of the non-categorical slots with other values seen for the same slot in other dialogues randomly. It enables us to create new dialogues from the available dialogues in an offline manner. However, augmentation in multi-turn dialogues can be challenging as changing a value for a slot may need some update in the rest of dialogue to have keep the altered dialogue still consistent. We carefully updated the whole dialogue with each slot's value replacement to maintain the integrity of the dialogue's content and annotations.

Augmentation for categorical slots was not possible since the dataset does not provide the unique position of the categorical slot values in the dialogue utterance. Also, we did not try the augmentation on multi-domain dialogues as switching between services makes it more challenging to maintain the consistency of the dialogue.

We augmented the data with 10x of the original data, but kept the number of training steps fixed to have a fair comparison by keeping the total amount of the computation the same. The results reported in Table \ref{data-augmentation} show the effectiveness of augmentation on both the SGD and SGD+ datasets for single domain dialogues. The experiments on the SGD dataset are done just for the seen services.

\begin{table}[!b]
\caption{\label{data-augmentation} FastSGT with and without data augmentation (Aug.) on seen single-domain dialogues of SGD and SGD+ datasets. All results are reported for dev/test sets.}

\centering
\resizebox{\columnwidth}{!}{
\begin{tabular}{lc|cccc}
 \textbf{Dataset}  &\textbf{Aug.} & \textbf{Intent Acc}&\textbf{Slot Req F1}&\textbf{Average GA} & \textbf{Joint GA} \\ \hline
 SGD & -- & 98.86/73.98 & 99.64/99.24 & 96.54/95.31 & 88.03/81.56 \\
 SGD & + & 98.74/73.97 & 99.59/99.31 & 96.70/96.31 & \textbf{88.66/83.12} \\ \hline
 SGD+ & -- & 95.36/95.67 & 99.52/99.59 & 95.57/95.44 & 83.97/84.31 \\
 SGD+ & + & 95.31/95.76 & 99.49/99.65 & 96.23/96.19 & \textbf{84.93/86.24} \\
\end{tabular}}
\end{table}

\subsection{Ablation Study}
We study and evaluate the effectiveness of different aspects of our model in this section. The performances of different variants of FastSGT are reported in Table \ref{ablation-study}. In each variant, one part of the model is disabled to show the effect of each feature of the model on the final performance. As it can be seen, both carry-over mechanisms are very effective and most of the improvements of our model is the result of this part of the model. The experiments show the effectiveness of the cross-service carry-over for multi-domain dialogues which was expected. We did not report the results of removing the cross-service carry-over for single-domain dialogues as it does not affect such dialogues.

The performance of the variant of the model without carry-over status or carry-over values for categorical slots is still good. The main reason is that while this variant of the model lacks these features it still can handle carry-over for non-categorical slots by predicting "active" status along with out-of-bound span prediction.

Experiments also show that the attention-based projections can help the performance of the model in terms of accuracy. It shows the effectiveness of exploiting all the encoded outputs instead of just the first output of the encoder by using self-attention mechanism.

\section{Conclusions}
In this paper we proposed an efficient and robust state tracker model for goal-oriented dialogue systems called FastSGT. In many cases in multi-turn dialogues, the value of a slot is not mentioned explicitly in the user utterance. To cope with this issue, the proposed model incorporates two carry-over procedures to retrieve the value of such slots from the previous system utterances or other services. Additionally, FastSGT model utilized multi-head attention mechanism in some of the decoders to have a better modeling of the encoder outputs.

We run several experiments and compared our model with the baseline model of the SGD dataset. The results indicate that our model has significantly higher accuracy, at the same time being efficient in terms of memory utilization and computation resources. In additional experiments, we studied the effectiveness of augmentation for our model and shown that data augmentation can improve the performance of the model. Finally, we also included ablation studies measuring the impact of different parts of the model on its performance.

\label{conclusion}

\bibliographystyle{ACM-Reference-Format}
\bibliography{sample-base}

\end{document}